\documentclass[12pt,twocolumn,twoside]{article}

\usepackage{titlesec}
\usepackage{fancyhdr}
\usepackage{authblk}
\usepackage{siunitx}
\usepackage{setspace,verbatim}
\usepackage{transparent}

\usepackage{xcolor,colortbl}
\usepackage{graphicx}
\usepackage{longtable}
\usepackage{booktabs}
\usepackage{multirow}
\usepackage[colorlinks, linkcolor=blue, anchorcolor=blue, citecolor=blue]{hyperref}
\usepackage{subcaption}
\usepackage{amsfonts}
\usepackage{amssymb}
\usepackage{amsmath}
\usepackage{lscape}
\usepackage{blindtext}
\usepackage{subcaption}
\usepackage{textcomp}
\usepackage{gensymb}
\usepackage{threeparttable}
\usepackage{booktabs, caption, makecell}
\usepackage{url}

\usepackage{breakurl}
\usepackage{hyperref}

\usepackage[top=1in,bottom=1in,left=1in,right=1in]{geometry}

\def\01{\ensuremath{0\mathord{-}1}}
\def\st{\mathop{\rm s.t.}}

\def\R{{\mathbb R}}

\usepackage[customcolors]{hf-tikz}

\tikzset{style green/.style={
    set fill color=green!50!lime!60,
    set border color=white,
  },
  style cyan/.style={
    set fill color=cyan!90!blue!60,
    set border color=white,
  },
  style orange/.style={
    set fill color=orange!80!red!60,
    set border color=white,
  },
  hor/.style={
    above left offset={-0.15,0.31},
    below right offset={0.15,-0.125},
    #1
  },
  ver/.style={
    above left offset={-0.1,0.3},
    below right offset={0.15,-0.15},
    #1
  }
}

\titleformat*{\section}{\normalsize\bfseries\sffamily}
\titleformat*{\subsection}{\small\bfseries\sffamily}
\titleformat*{\subsubsection}{\small\bfseries\sffamily}

\usepackage[font=small,labelfont=bf]{caption}

\newcommand\shorttitle{The ALAMO approach to machine learning}
\newcommand\authors{Wilson and Sahinidis}

\fancypagestyle{firststyle}
{
   \fancyhf{}

   \fancyfoot[L]{\footnotesize\scshape \copyright 2017. This manuscript version is made available under the CC-BY-NC-ND 4.0 license \url{http://creativecommons.org/licenses/by-nc-nd/4.0/}.  The formal publication of this article is available at \url{https://doi.org/10.1016/j.compchemeng.2017.02.010}.}
}

\title{\textbf{The ALAMO approach to machine learning}}
\author[1]{Zachary~T.~Wilson}
\author[1]{Nikolaos~V.~Sahinidis}
\affil[1]{Department of Chemical Engineering, Carnegie Mellon University, Pittsburgh, PA, USA}
\date{}

\usepackage[numbers,comma]{natbib}
\begin{document}

\begin{singlespacing}
\twocolumn[{%
%%% title
\maketitle
\thispagestyle{firststyle}

% Define bibstyle
\bibliographystyle{plainnat}

%\doublespacing
%\baselinestretch

%\section**{Abstract}
%\vspace{-5pt}
%\pagenumbering{gobble}
\footnotesize
\vspace{-1cm}
\begin{abstract}
ALAMO is a computational methodology for leaning algebraic functions from data. Given a data set, the approach begins by building a low-complexity, linear model composed of explicit non-linear transformations of the independent variables. Linear combinations of these non-linear transformations allow a linear model to better approximate complex behavior observed in real processes. The model is refined, as additional data are obtained in an adaptive fashion through error maximization sampling using derivative-free optimization. Models built using ALAMO can enforce constraints on the response variables to incorporate first-principles knowledge. The ability of ALAMO to generate simple and accurate models for a number of reaction problems is demonstrated. The error maximization sampling is compared with Latin hypercube designs to demonstrate its sampling efficiency. ALAMO's constrained regression methodology is used to further refine concentration models, resulting in models that perform better on validation data and satisfy upper and lower bounds placed on model outputs.
\end{abstract}


\vspace*{0.8cm}

}]

\footnotesize
%\newpage

\section{Introduction}
\label{sec:intro}

The use of data science and machine learning techniques has become pervasive across disciplines of science and engineering to address problems associated with increasingly large and complex data sets. The abundance of data available to analysts comes from a number of sources, including the prevalence of hard and soft sensors in process systems~\citep{kgs09}, the increased computational power available to perform complex simulations and store vast amounts of data, and the availability of modern social networking platforms to collect massive amounts of data from observational studies~\citep{nya05,wzwd14}. The problems presented to scientists and engineers often involve the optimization, prediction, control, and design of the system or process that generates a data set~\citep{ks00ejor,bcsp16}. Understanding the nature of the problem is critical in choosing an effective approach to accurately model the system at hand. Many problems in machine learning, including speech recognition, face recognition, and self-driving cars, are described by data sets that can have millions of samples or more and as many features~\citep{cml14,n09}. Accordingly, difficulty encountered in any analysis to follow can, in part, be ascribed to the pure volume of data~\citep{htf13,j09}. Complexity is also inherent in trying to understand the system or process that generates data. In many cases, these systems may be well suited to the accurate prediction of an output given an input, but can be poorly suited to tasks involving optimization or design~\citep{agpk03,fga05}.

In order to accurately represent a system, any candidate model for the system must have the expressive power to represent the behavior of the observed response. The field of representation learning, sometimes called deep learning, is specifically concerned with generating candidate models that are fully capable of representing nonlinear relationships between inputs and outputs, modeling multi-scale effects, and even learning significant features in an unsupervised fashion; while still being able to train the prospective model to some training data efficiently~\citep{bcv13,lbh15}. Recent efforts by companies including Google and Microsoft are specifically aimed at making break-through progress in the field of deep learning~\citep{deng13}. Specifically, deep learning refers to a number of learning techniques that typically address \emph{unsupervised} learning problems using different flavors of neural networks that contain many nodes in structured hidden layers. The learning problems being unsupervised simply means that the training data is not associated with any response vector. Instead, the goal of deep learning is to learn intrinsic relationships in the data set. This affords the opportunity to learn classes or aspects that are intrinsic to the data, and to make predictions about future data sets. As a result, deep learning methods can be generalized to both train a model using response vectors as well as predict a response vector in the absence of one. Unsupervised learning problems almost exclusively consist of classification problems. Deep learning techniques have been made popular to a large audience due to their recent documented successes. However, their powerful unsupervised learning techniques can also be applied to other situations, such as fault detection in a chemical plant~\citep{kjk07}.

Supervised learning techniques rely on a target vector of responses $z$ to train a candidate model. When this response is an integer, the problem is called classification. It is common to reformulate any classification problems, regardless of the number of classes, to a number of binary classification problems. The remainder of this paper will focus on the regression setting, where $z \in \R^N$. Within this domain of supervised regression problems, approaches can be divided into either parametric or non-parametric regression techniques. \emph{Non-parametric regression} methods include a number of different methodologies, including artificial neural networks~\citep{h09,vapnik03}, support vector machines~\citep{htf13,ss01}, and $k$-nearest neighbors~\citep{a92} that function by placing different weights on each sample $i=1, \ldots, N$, and predicting responses based on an appropriate averaging scheme. \emph{Parametric regression} applies a different weight to each selected feature $j = 1, \ldots, p \le k$, and predicts the responses as a function of the features. This function is a linear combination of various parameterizations of the inputs in the case of linear regression.

Although many inference techniques have recently been developed in the context of large data sets, they are routinely used for situations in which there is a dearth of both data points and number of features. In chemical kinetic parameter estimation, acquisition of data can be expensive~\citep{h77,bfm10}. As a result, parameter estimations are made in an attempt to identify unique parameter values from a minimum number of experimental data points. This places a high premium on the identification of simple models that do not overfit training data, the ability to incorporate information from first principles to bolster the model quality, and the intelligent acquisition of new data points if available. Different techniques have been developed to address these issues in the context of learning an appropriate regression or classification model.

\cite{csm:14} recently developed the ALAMO approach to machine learning, motivated by the need to obtain simple models from experimental and simulation data.  Soon thereafter, the approach was extended to combine data-driven and theory-driven model building~\citep{csm:15}.  The primary purpose of this paper is to review, evaluate, and illustrate the ALAMO methodology and contrast it to existing techniques in the machine learning literature.  Section~\ref{sec:review} presents a review of machine learning approaches to model building.  In Section~\ref{sec:alamo}, we discuss the core concepts of the ALAMO methodology. Subsection~\ref{sec:mip} describes the application of this methodology to learning simple models from a fixed data set. In Subsection~\ref{sec:samp}, ALAMO's adaptive sampling methodology is reviewed. ALAMO's unique ability to enforce physical constraints on the response of the regression model is explored in Subsection~\ref{sec:conreg}. An illustrative example is presented in Section~\ref{sec:ex} to provide insights to the algorithms presented.  Finally, extensive computations are presented on a set of 150 problems in Section~\ref{sec:com}, followed by conclusions in Section~\ref{concs}.

\section{Review of current model building methodologies}
\label{sec:review}

In a typical linear least squares regression setting, a system's responses $z_i , i=1,\ldots,N$, are approximated through linear combinations $\hat{z} = X \beta$ of a design matrix $X$, where $X_i = (x_{i1}, \ldots, x_{ip}), x_{ij} \in \R$ are the regressors and $\beta \in \R^p$ are the regression coefficients. Ordinary least squares (OLS) estimates for the regression coefficients can be obtained in closed form as $\hat{\beta}_{ols} = \left( X^tX \right)^{-1} X^t z$, and are obtained by minimizing the sum of squared residuals,
\begin{equation*}
\text{SSR}=\sum_{i=1}^N \left( z_i- \sum_{j=1}^p \beta_j x_{ij} \right) ^2
\end{equation*}

Linear models often incorporate an increasing number of regressors, either through acquisition of new independent variables or transformations of existing ones. As the basis set of regressors increases in size, the model will begin to describe noise and other aspects of the training data, a phenomenon commonly referred to as overfitting. Models that overfit data have a low bias, resulting from the increased degrees of freedom in the regression model, but high variance, meaning the predicted values can change dramatically with small changes in the training data~\citep{htf13}. Balancing the bias-variance trade-off and preventing overfitting in data-driven models is of fundamental importance across modeling techniques in order to produce models that will generalize to unseen data~\citep{sl02}.

Avoiding overfitting is facilitated in practice by eliminating, or diminishing the effects of, superfluous regressors. A common method for accomplishing this is the minimizing of an objective function composed of the sum of the $SSR$ on a training set, and some $q$-norm of the regression coefficients, which quantifies the complexity of the linear model.
\begin{align*}
& \text{BSS}=SSR + C \| \beta \|_q \\
& \| \beta \|_q = \left( \sum_{j=1}^p \hat{\beta}_j ^q \right) ^{1/q}
\end{align*}
If $q=0$, this norm is called the $l_0$ (pseudo)norm, and is simply the number of nonzero regression coefficients $\| \beta \|_0 = \sum_{j=1}^p I ( \hat{\beta}_j \ne 0 )$, where $I()$ is the indicator function. The value of $q$ determines a great deal about the properties of this problem; $q$ values of 0, 1, and 2 are referred to as best subset selection, the lasso, and ridge regression respectively, and combinations of $l_1$ and $l_2$ norms is refereed to as elastic net regression~\citep{fht10:lassoimp,t96:lasso,htf13}. The parameter $C$ quantifies the relative trade-off between the complexity of a model and its error on a training set. Although minimization of the best subset problem has shown to produce more parsimonious models than the alternative problems, the use of the $l_0$ norm has been historically avoided due to the computational complexity associated with this combinatorial problem. However, integer optimization algorithmic advances combined with improved hardware have recently allowed for the best subset problem to be practically solved for problems with hundreds or thousands of regressors.

Accurate models can significantly reduce the expense associated with the design, optimization, and control of complex systems. For the purposes of developing accurate system models, standard multiple linear regression using the inputs to the system may be insufficient to develop accurate predictive models. A linear combination of the original inputs into the system can be inadequate to describe nonlinear relationships endemic in the underlying process. In this case, it is common to rely upon non-parametric techniques. These non-parametric techniques can build highly accurate representations of complicated nonlinear system responses~\citep{cv95,ss01}; however, this accuracy comes at the expense of model complexity and a lack of interpretability. The larger problem is that of designing a hypothesis space of all possible model forms and parameter values that are capable of accurately representing the observed response. This fundamental question is intricately tied to the field of representation learning, although some efforts to identify optimal nonlinear parametric transformations have been made~\citep{bf85}.

The goal of building an inference-based model may not only be prediction accuracy, but also construction of a model that is informative of the underlying process. It is common for a significant amount of time and effort to be spent on the development of a nonlinear regression model from first principles. In the absence of insight or experience with a given domain, a standard set of parameterizations may be used as is the case in the standard polynomial regression to construct a new set of features. Although the new set of features will undoubtedly exhibit a high degree of correlation and may still not contain an adequate representation of the underlying process, performing model selection on the new set of features can produce models that are simple, interpretable, and highly accurate~\citep{k94,csm:14}. Even if predictive accuracy is of the utmost importance, a correctly specified parametric model will have a smaller variance than an optimal non-parametric model due to their flexibility~\citep{c16}. The expanded set of regressors can accurately represent nonlinear system responses, but will be prone to overfit the training data. This motivates the need to select the best subset of regressors in a multiple linear regression~\citep{m02}.

In linear models, preventing overfitting is often associated with finding the best subset of regressors. Best subset selection, also known as model selection or feature selection, has long been an active field in statistics~\citep{h76} and has received recent attention in the fields of machine learning and optimization~\citep{bs07,lm07,ky09,csm:14,mt15,bk15,pk13ss}. Exhaustive search algorithms, the most popular of which is Leaps-and-Bounds~\citep{fw74}, were first proposed to address the subset selection problem. A major advantage of these exhaustive methods is that they can provide a number of viable models, all of which could perform adequately on unseen data. This small selection of models can then be validated on a test set, or through cross validation.

Due to the combinatorial nature of the search for the best subset, which is known to be NP-hard~\citep{ak98}, exhaustive search methods do not scale well with the size of the problem. A popular implementation of this algorithm~\citep{leapsR} is limited to only 32 regressors. This computational complexity results from exhaustive search algorithms as well as other embedded techniques, in which model selection and parameter estimation are performed simultaneously~\citep{ge03}.  As a result, these algorithms are often passed over in favor of filter or wrapper methods. \emph{Filter methods} involve a one-pass ranking of all features, and the discarding of any features that do not satisfy the chosen filter. \emph{Wrapper methods} actively add and remove possibilities from the candidate model, often in a greedy fashion~\citep{jkp94,kj97}. Many wrapper methods can be generalized beyond the context of best subset selection in multiple linear regression, but are also directly applicable to this problem. Popular techniques include forward selection, backward elimination, and combinations of the two which fall under the umbrella term stepwise regression~\citep{ba77,m02}. Stepwise heuristics for finding the best subset involve adding or removing regressors in an iterative fashion until a termination criterion is met. These stepwise heuristics typically produce good models, but offer no guarantee of optimality. Moreover, there exist a number of examples in the literature illustrating other shortcomings of these heuristics~\citep{m02}. Filter methods can also be applied to a number of model selection techniques including best subset selection. These methods are used to screen the features, routinely as part of a pre-processing procedure.

Regularization techniques also combat overfitting in linear models, and can facilitate model selection when regression coefficients are driven to zero (e.g.,~\cite{t96:lasso}). These techniques rely on solving convex continuous optimization problems where the objective is a trade-off between error on the training set, and the magnitude of the regression coefficients:
\begin{equation}
\label{eq:reg}
\hat{\beta} = \text{argmin}_\beta \frac{1}{2} \| z - X \beta \|_2^2 + \lambda \alpha \|\beta \|_1 + (1-\alpha) \lambda \| \beta \|_2
\end{equation}
where the parameter $\alpha$ determines the trade-off between the $l_1$ penalty and the $l_2$ penalty. If $\alpha=1$, then Equation~(\ref{eq:reg}) is equivalent to the lasso problem~\citep{t96:lasso}, and will facilitate subset selection as the $l_1$ penalty will drive regression coefficients toward zero. If $\alpha=0$, the problem is equivalent to ridge regression~\citep{hk70:ridge}. For any value of $\alpha$ between 0 and 1, the problem is termed elastic net regression. Despite the fact that regularization does facilitate subset selection, Model~(\ref{eq:reg}) is not equivalent to the problem of best subset selection. Although regularization techniques do retain computational advantages, the use of mixed-integer programming (MIP) to facilitate subset selection has been shown to produce models that are smaller, and more accurate, than other techniques including the lasso~\citep{csm:14}.

A number of model fitness metrics have been developed from information theory and statistical theory to balance the bias-variance trade-off. Model fitness metrics that will be investigated in this paper include the corrected Akaike's information criterion (AICc)~\citep{a74}, Mallows' Cp (Cp)~\citep{m73:cp}, the Bayesian information criterion (BIC)~\citep{s78,w96bicwilcox}, the Hannan-Quinn information criterion (HQIC) \citep{qh79}, the risk inflation criterion (RIC)~\citep{fg94}, and mean squared error (MSE)~\citep{pk13ss}. Different philosophies were followed in the derivation of each metric. For example, AIC seeks to minimize the Kullback-Leibler divergence between an underlying true distribution and the estimate from a candidate model, BIC attempts to maximize the posterior model probability, and Cp seeks to minimize the mean square error of prediction. The differing perspectives utilized in the derivations result in the identification of different best subsets of regressors. The statistical motivation of some model fitness metrics, specifically utilizing Kullback-Leibler divergence as a means of model selection, has been thoroughly investigated in the literature~\citep{ba01,ba02,m02}. All metrics account directly for the inclusion of an explanatory variable when quantifying model complexity, and this makes them all highly amenable to MIP formulations. Standard assumptions made for linear regression, namely the residuals being normally distributed with zero mean and a specified variance $\sigma^2$, result in similar functional forms across fitness metrics. These metrics are defined as follows.

%\begin{equation*}
%FM( \beta, y)
%\end{equation*}

%Integer programming formulations for minimizing $\text{C}_P$~\citep{mt15},  MSE~\citep{pk13ss}, and AIC~\citep{e79,slw98} have been investigated in the literature. However, these investigations focus on a handful of algorithms for minimizing one fitness metric, and do not provide a broad comparison with other fitness metrics or alternative subset selection methods. This makes a direct comparison between metrics and techniques difficult to ascertain from the current literature.
%
%\begin{singlespace}
\begin{equation}
\label{eq:aic}
\text{AIC}_c = N \ \log \left( \frac{1}{N} \sum_{i=1}^N (z_i - \textbf{X}_i \beta )^2 \right) +2p + \frac{2p(p+1)}{N-p-1}
\end{equation}
\begin{equation}
\label{eq:hqic}
\text{HQIC} = N \ \log \left( \frac{1}{N} \sum_{i=1}^N (z_i - \textbf{X}_i \beta )^2 \right) +2p\log(\log(N))
\end{equation}
\begin{equation}
\label{eq:mse}
\text{MSE} = \frac{\sum_{i=1}^N (z_i - \textbf{X}_i \beta )^2}{N-p-1}
\end{equation}
\begin{equation}
\label{eq:cp}
\text{C}_P = \frac{\sum_{i=1}^N (z_i - \textbf{X}_i \beta )^2}{\hat{\sigma}^2}+2p-N
\end{equation}
\begin{equation}
\label{eq:bic}
\text{BIC} = \frac{\sum_{i=1}^N (z_i - \textbf{X}_i \beta )^2}{\hat{\sigma}^2}+p \ \log(N)
\end{equation}
\begin{equation}
\label{eq:ric}
\text{RIC} = \frac{\sum_{i=1}^N (z_i - \textbf{X}_i \beta )^2}{\hat{\sigma}^2}+2p \ \log(k)
\end{equation}
%\end{singlespace}
%
In all model fitness metrics, $N$ is the number of data points in the training set, $p\le k$ denotes the number of nonzero regression coefficients, where $k$ is the total number of regression coefficients. The parameter $\hat{\sigma^2} = \frac{SSR_{ols}}{N-1}$ is an estimation of the residual variance, and is usually obtained from the mean square error of the full $k$-term model, which is an unbiased estimator. All metrics utilize the SSR, or $l_2$ error, to quantify fit on a training set, and penalize this fit directly based on the number of nonzero regression coefficients. Additionally, three of the six metrics are integer convex quadratic functions~(\ref{eq:cp})--(\ref{eq:ric}) and can be optimized directly through mixed-integer quadratic (MIQP) formulations. Direct optimization of the remaining metrics~(\ref{eq:aic})--(\ref{eq:mse}) would require mixed-integer nonlinear programming (MINLP). The derivation of the functional form of Bayes information criterion is obtained by utilizing the likelihood function $\hat{L}=\frac{1}{\sqrt{2 \pi \sigma^2}} \exp \frac{- (z - \hat{z}) ^2}{2 \sigma^2}$ obtained by assuming the residuals of the model are normally distributed with zero mean, and variance estimated by $\hat{\sigma} = \frac{SSR}{N-1}$.

\section{ALAMO}
\label{sec:alamo}
%use this section to summarize ALAMO's core components.

\begin{figure*}
\centering
%\figuretitle{ALAMO algorithmic flowchart}
\includegraphics[scale=0.9]{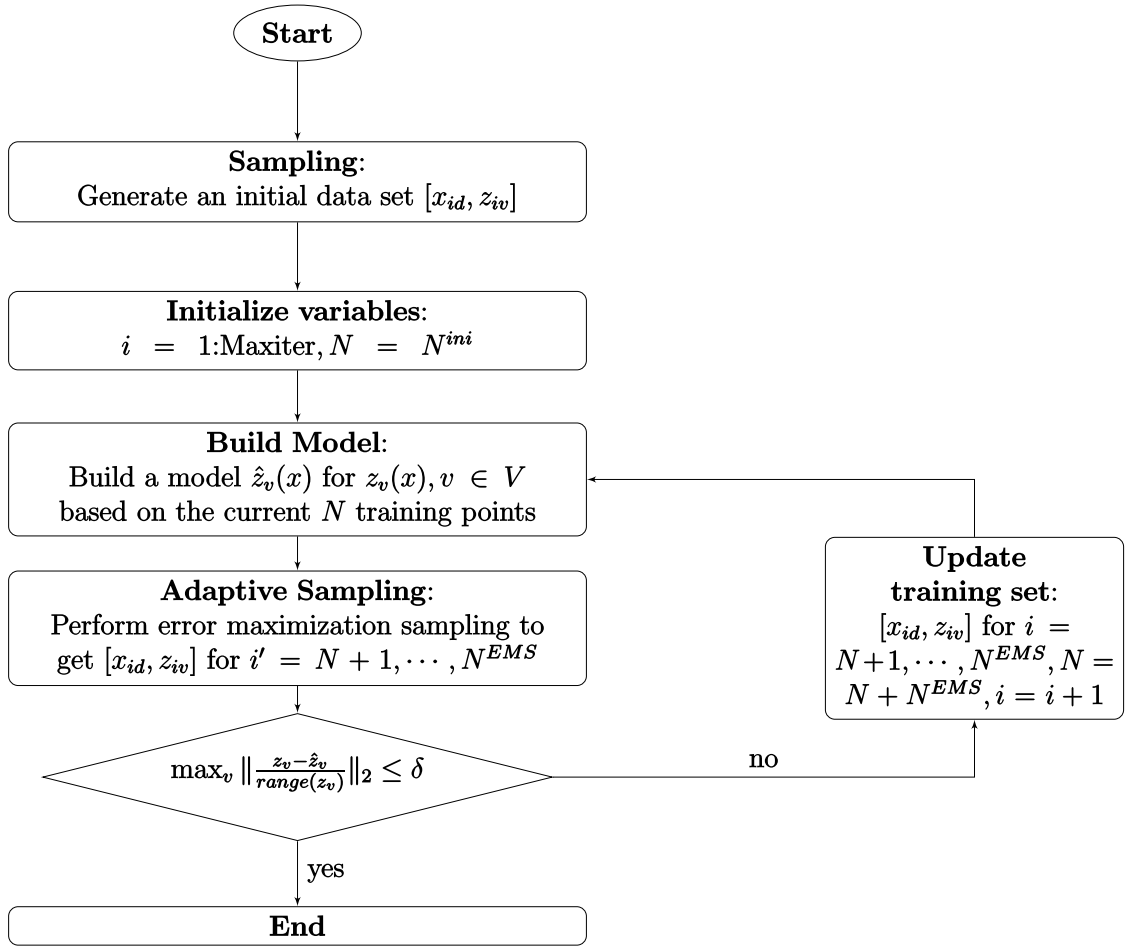}
\caption{Flowchart of model building algorithm utilized by ALAMO}
\label{fig:alamo}
\end{figure*}

ALAMO is a regression and classification model learning methodology that builds simple, accurate surrogate models from experiments, simulations, or any other source of data using a minimal set of sample points. At the heart of ALAMO, is an integer-programming-based best subset technique that considers a large number of explicit transformations of the original input variables $x$. The model is then tested and improved using derivative-free optimization solvers that sample promising points in an adaptive fashion. The model building and adaptive sampling features constitute the core of ALAMO as seen in Figure~\ref{fig:alamo}. The process begins with an initial data set of $N^{ini}$ points. A surrogate model is then built, and an adaptive sampling methodology referred to as error maximization sampling (EMS) is utilized to identify the next $i^{'} = N+1 \cdots N^{ems}$ points. If the selected model satisfies some $\delta$ tolerance on the maximum observed error, then the algorithm is allowed to terminate; otherwise, the points identified by EMS are appended to the training set.

This section contains a summary of the core components of the computational implementation of the ALAMO methodology. In the usual setting, ALAMO can be cast into the domain of linear parametric regression, where the columns of the design matrix are composed of explicit parameterizations of the original inputs into the system.

Any parametric regression methodology relies on explicit parameterizations of the system inputs to accurately describe any nonlinear behavior observed in the response data. A typical set of nonlinear transformations may be comprised of a number of transformations, including monomial, binomial, ratio, exponential, logarithmic, and trigonometric. Additionally, more complicated nonlinear functional forms such as a sigmoid, Arrhenius, and Gaussian relationships can be accommodated if the regression parameter in the nonlinear relationship is fixed and made to serve the role of a hyper parameter. Although this paper will focus on ALAMO's ability to produce parametric regression models, the addition of any distance function such as the Gaussian function will extend this methodology into the realm of non-parametric regression.

\subsection{Learning simple surrogates}
\label{sec:mip}

In this section, a MIP formulation for the optimization of the model fitness metrics discussed to facilitate best subset selection in linear regression is examined. As in the usual regression setting described before, we begin with a set of $N$ training points; where each data point $(i=1, \ldots, N)$ contains a set of input data $x_{id}, d \in D$, and a set of responses $z_{iv}, v \in V$. In the interest of simplifying the presentation, we will restrict our attention to the problem of a single response vector. A number of nonlinear transformations will be performed in order to populate a regression design matrix, $X_{ij}, j=1,\ldots,k$. %As mentioned earlier, non-parametric functional forms such as distance functions could be included to the basis set to extend the representation power of a candidate model.

In order to balance the bias-variance trade-off endemic to all data-driven modeling techniques, a MIP formulation will be used to optimize a model fitness metric. Despite nonlinearities present in some fitness metrics, they can all be optimized by parametrically minimizing the SSR subject to a maximum number of nonzero regression coefficients. In particular, any model fitness metric can be optimized through a series of cardinality constrained mixed-integer quadratic programs (CCMIQP) defined as follows:
\begin{equation}
\min_{r = 1, \ldots ,p} FM( r ) \label{eq:cc}
\end{equation}
where
\begin{equation*}
\begin{aligned}
FM( r ) = & \min \; \sum_{i=1}^{N} \left(z_i - X_i \beta \right)^2 + C(r) \\
& \st \; \sum_{j = 1}^p y_j \le r \\
& \quad \; -M y_j \le \beta_j \le M y_j, \quad j = 1 ,\ldots ,p  \\
& \quad \; y_j \in \{0,1\}, \quad j=1,\ldots,p   \\
\end{aligned}
\end{equation*}
Above, $FM$ is any model fitness metric~(\ref{eq:aic})--(\ref{eq:ric}), and $C(r)$ is a metric-dependent complexity penalty; $C(r)$ is constant for a specified model cardinality $r$. A cardinality constraint on the binary variables, $y_j$, ensures no more than $r$ regressors are included in the model. The binary variables are used in conjunction with big-$M$ constraints to model the inclusion or exclusion of a regressor. The value of the constant, $M$, in these big-$M$ constraints is found using logic borrowed from the lasso, ($M=\sum_{j=1}^{p} | \hat{\beta}_{j,ols} |$).

Although Formulation~(\ref{eq:cc}) can optimize all six model fitness metrics of interest, three metrics~(\ref{eq:cp})--(\ref{eq:ric}) will admit the following convex MIQP formulation for their direct optimization defined in the following formulation:
\begin{equation}
\left.
\begin{aligned}
\min & \quad FM  \\
\st &  \quad  -M y_j \le \beta_j \le M y_j, \quad j=1,\ldots,p \\
& \quad y_j \in \{0,1\}, \quad j=1,\ldots,p  \\
\end{aligned}
\right\rbrace \label{eq:bss}
\end{equation}

Formulation~(\ref{eq:bss}) considers all $2^k$ possible regression models, including the null model, while Formulation~(\ref{eq:cc}) considers only $\left( \frac{k}{r} \right)$ models under the cardinality constraint. These two formulations are analogous, although by no means equivalent, to the penalized as well as constrained versions of the lasso problem. The constrained version of the lasso $\{ \min_{\beta} SSR | \|\beta\|_1 \le T \}$, for some parameter $T$, may be more interpretable, as any $T < \| \hat{\beta}_{ols} \|_1$ will cause shrinkage in the regression coefficients. However, the use of the penalized version of the lasso $\{ \min_{\beta} SSR + C \|\beta\|_1 \}$, for some trade-off parameter $C$, is more common in practice due to a large number of computationally efficient implementations for solving this problem~\citep{fht10:lassoimp}. The issue of interpretability is easily avoided, as a maximum value for the parameter $C$ where all regression coefficients are zero can be found directly from the KKT conditions of the lasso problem~\cite{fht10:lassoimp}. It is common to start with this null solution, and find subsequent solutions for lower values of $C$ utilizing the previous solution, a process called warm-starting. Extending this back to the case of best subset selection, the CCMIQP Formulation~(\ref{eq:cc}) affords direct control on how many regression coefficients are nonzero, but may prove computationally inefficient if a large number of regressors are to be included in the optimal model. Likewise, the convex MIQP Formulation~(\ref{eq:bss}) may provide a computationally expedite solution of the same problem. However, unlike the penalized lasso, the fit-complexity trade-off found in~(\ref{eq:bss}) is informed from theory instead of exhaustively solved for using warm-starting.

\subsection{Adaptive Sampling}
\label{sec:samp}

Adaptive sampling, sometimes called active learning, refers to a means by which information about the underlying response surface can be acquired by querying the system at desired input levels. The ability to identify a data set that will perfectly train a model, or an optimal design of experiments (DOE), a priori does not exist. Instead, DOE methodologies could try to fill the space of the input variables in an optimal fashion. These techniques are referred to as single pass techniques and include factorial designs~\citep{simpson01}, Latin hypercubes~\citep{mbw00,fl03}, and orthogonal arrays~\citep{hk12}. They generate a set of data points in the domain of the input variable, evaluate these points, and then move on to train the model. These methods are appealing in that they are often easy to implement and computationally efficient. However, the data set these methods produce may not be optimally suited to training a model. Since data is obtained in such a way that the input domain is filled, specific problem areas in the domain may be underrepresented, making the resultant models lack fidelity. Iterative techniques make use of the current model and data set to locate promising new sampling points. Exploration-based iterative methods seek to fill in the space of the input domain in a fashion similar to single pass methods. Exploitation-based methods utilize models in an adaptive fashion to locate data points near areas that are difficult to model. These may include regions of sharp nonlinearities or discontinuities.

The goal of adaptive sampling is to optimize the objective
\begin{equation}
\begin{aligned}
\max_{x^l \le x \le x^u} & \left( \frac{z(x)-\hat{z}(x)}{z(x)} \right) ^2 \\
\end{aligned}
\label{eq:ems}
\end{equation}
This formulation represents an error maximization sampling (EMS) approach. The algebraic form of $\hat{z}(x)$ is obtained through the use of ALAMO, or another surrogate modeling technique, but the algebraic form of $z(x)$ is not known. The true black-box value of the objective is not known; therefore, a class of algorithms utilizing derivative-free optimization (DFO) must be used. A comparative study~\citep{rs13} has investigated the use of a number of DFO solvers and discovered that SNOBFIT~\citep{hn08} is currently a highly effective DFO solver for minimizing problems such as~\ref{eq:ems}. The derivative-free solvers operate by estimating the error at new candidate points. If the estimated error indicates that the new point is in a region of model mismatch, then the point is added to the training set and a new model is built. The points to be added to the training set can also serve as a conservative estimate on the true value of~(\ref{eq:ems}). If this true error estimate is above the specified tolerance of $\delta$ the model is retrained using the updated data set as seen in Figure~\ref{fig:alamo}. If the true error estimate is below the tolerance, then the proposed model has converged on the final surrogate model of $z(x)$.

\subsection{Constrained Regression}
\label{sec:conreg}

Data-driven models are routinely used when domain specific knowledge is available a priori that can be utilized to refine potential models. A fundamental form of restriction is composed of an upper and lower bound on the regression variables $\beta$. This leads to the usual constrained regression problem
\begin{equation}
\begin{aligned}
\min_{\beta \in A} \; g ( \beta ; x, z ) \\
\end{aligned}
\end{equation}
This model determines the regression parameters $\beta \in \R^p$ that minimize a given loss function over an original set of regression constraints $A$. We wish to enforce an additional set of constraints
\begin{equation}
\label{eq:fr}
\Omega(\chi) := \left\lbrace \beta \in \R^m:f[x,\hat{z}(x;\beta)] \le 0, x\in\chi \right\rbrace
\end{equation}
where the function $f$ is a constraint in the space of the modeled response(s) $\hat{z}(x)$ and the transformed predictors $x$, and $\chi$ is a nonempty subset of $\R^n$. The generality of Equation~(\ref{eq:fr}) extends beyond the ALAMO methodology. This equation can be used to reduce the feasible region for any regression analysis, including least squares, regularization techniques, best subset methods, and other nonparametric regression techniques. The general regression problem is now reformulated as
\begin{equation}
\label{eq:cr}
\begin{aligned}
\min_{\beta \in A\cap \Omega(\chi)} \; g ( \beta ; x, z ) \\
\end{aligned}
\end{equation}
Previous work has employed a priori knowledge to reveal relationships between regression coefficients~\citep{rao65,b74,kk11}. This may be of particular use in hierarchical regression tasks, where explicit rankings or importance of features may be known a priori. Inequality relationships between regression parameters have been used both in linear and nonlinear regression in engineering~\citep{gm99}, statistical~\citep{jt66}, as well as economic applications~\citep{t82}. The sum of this previous work attempts to relate the regression coefficients through relationships obtained a priori. Instead, a more desirable constraint is one placed on the estimated response, that can be guaranteed across a domain of interest.
The desire to constrain the response of the regression models leads to a complication in the realization that Eq.~\ref{eq:fr} is valid over the full domain of the inputs; requiring this equation to be enforced at infinitely many points, i.e., $\forall x \in \chi$. In the larger regression problem, this results in a problem with countably many ($p$) regression variables, but infinitely many possible constraints for every equation $f$ that is applied to possible models. Semi-infinite programming (SIP) problems are optimization models that can be used to consider problems with a finite number of variables, but an infinite number of constraints. The core component of SIP involves solving $\max_{x \in \chi} f [x, \hat{z}(x ; \beta)]$ to locate the point in the domain at which the maximum violation in the current surrogate model and the applied constraint occurs. This subproblem is significant because $\beta \in \Omega(\chi)$ if and only if $\max_{x \in \chi} f [x, \hat{z}(x ; \beta)] \le 0$. Therefore, this problem is in general nonlinear and nonconvex, thus requiring global optimization algorithms for solution.

Constraining a regression model's output in this fashion allows a modeler to: (a) restrict the bounds of the response, either individually or simultaneously restricting upper and lower bounds, (b) restrict the derivatives of the response to enforce model convexity or a monotonic relationship, and (c) enforce bounds over a domain outside of the training domain. Although regression-based models should not be used for extrapolation outside of the training domain, enforcing bounds on the response and certifying that a selected model satisfies these bounds can allow for a safer alternative to blind extrapolation. For a further discussion of ALAMO's constrained regression, see \cite{csm:15}.

\section{Illustrative example}
\label{sec:ex}

\begin{figure*}
\centering
\includegraphics[scale=0.5]{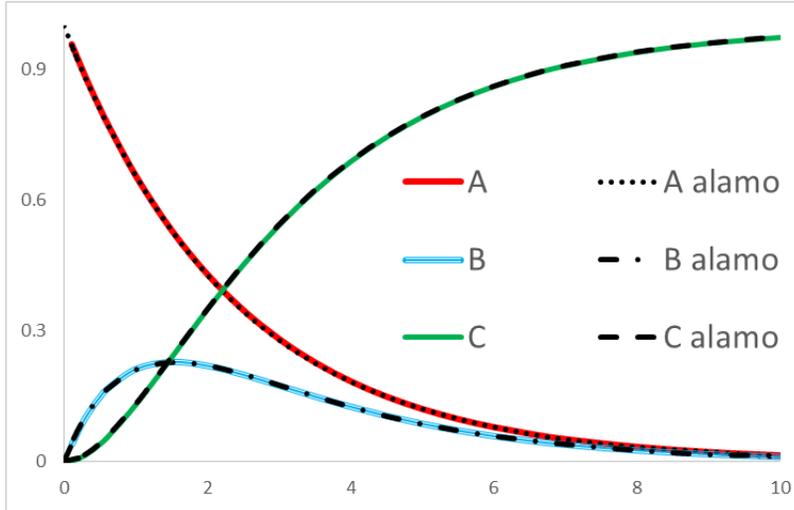}
\caption{Concentration profiles of all three components for the two reactions in series}
\label{fig:conc}
\end{figure*}

A detailed look into one problem is provided to enable a more interpretable perspective on the models generated by ALAMO, and in order to better facilitate the broader comparisons drawn later in the paper. There are two reactions in series $A \xrightarrow{k_a} B \xrightarrow{k_b} C$. We assume that the true system is governed by the following set of differential equations describing the transient mass balances of isothermal batch reactors:
\begin{equation*}
\begin{aligned}
& \frac{dA}{dt} = - k_1 A \\
& \frac{dB}{dt} = k_1 A - k_2 B\\
& \frac{dC}{dt} = k_2 B \\
\end{aligned}
\end{equation*}
The true nonlinear functional forms of the concentration profiles of the three species, with associated kinetic rate constants $k_1 = 0.42$ and $k_2 = 0.97$, and initial conditions $ A_0 = 1$, $B_0 = 0$, $C_0 = 0$ are as follows:
\begin{equation*}
\begin{aligned}
& A = \exp (-0.42 t) \\
& B = \frac{0.42}{0.97-0.42}*(\exp(-0.42 t)-\exp(-0.97 t)) \\
& C = 1 - A - B \\
\end{aligned}
\end{equation*}
Three concentration profiles are generated from ALAMO using data corresponding to a LHS design of experiments of 20 data points. The three corresponding models generated by ALAMO, without utilizing the constrained regression feature, are as follows:
\begin{equation*}
\begin{aligned}
& A_{\rm ALAMO} = 0.44 t + 0.00021 * t^{0.50} - 0.26 t^2 \\
& \qquad \qquad + 0.3 * t^{2.5} - 0.24  * t^3 + 0.101 t^{3.5}\\
& \qquad \qquad - 0.021 * t^4 + 0.002 t^{4.5} \\
& B_{\rm ALAMO} = 0.44 t - 0.48 t^2 + 0.3 * t^{2.5} - 0.08 t^{3} \\
& \qquad \qquad  + 0.008 t^{3.5} \\
& C_{\rm ALAMO} = 0.00021 * t^{0.50} + 0.22 * t^2 - 0.16 * t^3\\
& \qquad \qquad + 0.093 t^{3.5} - 0.021 * t^4 + 0.002 t^{4.5}
\end{aligned}
\end{equation*}
The basis set utilized for this experiment is $X_j \in [t^{\pm[0.5,1,1.5,2,2.5,3,3.5,4,4.5,5]},$ $ \log(t),$ $ \exp(t), 1]$. As seen in Figure~\ref{fig:conc}, this basis set is sufficient to obtain a perfect representation of all three concentration profiles.

In later comparisons, the time domain is sampled starting at later, nonzero, initial points. This is done for two primary reasons: first, it is often impractical to obtain concentration values from early process times due to time delays in the analytical process; secondly, the utility of constrained regression is more clearly demonstrated on a training set removed from 0, as the tendency of polynomial regression functions will be to have a sharp change in convexity in a region close, but outside, of the training set.

\section{Computational experiments}
\label{sec:com}

The computational comparisons presented in this section are intended to demonstrate the following:
\begin{itemize}
 \item The error maximization sampling (EMS) methodology is able to generate regression based models of a higher quality than those obtained from LHS.
 \item Constrained regression can be utilized in order to extend the range of regression-based models beyond the domain of the training data.
 \item For a given set of non-linear transformation, extending the range of the model comes at the expense of model accuracy on the training set; however, this problem is mitigated by the large number of candidate models available to ALAMO in the model building process.
\end{itemize}

In the following, the utility of the ALAMO methodology is demonstrated across a number of chemical reaction modeling problems. Data sets are generated from one of two possible reaction schemes. The concentration of three species is modeled from a batch reactor with two first-order reactions in series:
\begin{equation*}
A \xrightarrow{k_a} B \xrightarrow{k_b} C
\end{equation*}
as well as a batch reactor with two competing first-order reactions:
\begin{equation*}
\begin{aligned}
& A \xrightarrow{k_a} B \\
& A \xrightarrow{k_b} C
\end{aligned}
\end{equation*}
where the concentration of all species A, B, and C is a function of batch time $t$, where $0.6 \le t \le 10$, with the four associated kinetic rate constants $k_a $, where $ 0.4 \le k_a \le 3$ and $k_b$ , where $0.9 \le k_b \le 2.1 $ such that $k_a \ne k_b$. Five values are chosen for each parameter value via a uniform distribution and are exhaustively paired to yield 25 unique reactive systems. Boundary conditions for the reactors are available in the form of initial concentrations: $[A]_0 = 1 , [B]_0 = 0 , \text{and} [C]_0 = 0$. Training sets were sampled in one of two ways: using a Latin hypercube design of experiments for a specified number of points, or starting ALAMO with one random point over the domain $t_o \in [ 0.6 , 10 ]$ and allowing EMS to sample until a tolerance $\delta$ was obtained. In each problem, nonlinear transformations were applied to the input variable $t$ in order to create a basis set of 13 features, $X_j \in [t^{\pm[0.5,1,2,3,4]}, \log(t), \exp(t), 1]$, from which a surrogate model is generated. Details of the individual experiments are outlined in the appropriate subsection. As was demonstrated in~\ref{sec:ex} the simple kinetic schemes represented by the two reactions in series and parallel comprise a wide range of nonlinear behaviors to  be represented with a simple set of nonlinear transformations of the single input variable $t$.

In order to facilitate a comparison between the model-building techniques detailed in the remainder of the paper, a basis for the comparison will be established. The model fitness metric Bayesian Information Criterion (BIC) in Eq.~\ref{eq:bic} will be used as the objective function of all subset selection problems. The model identified as optimal by this fitness metric will first be compared in terms of its performance on a validation set, quantified through the root mean squared error ($RMSE_{val}$). All validation data will consist of 1000 randomly chosen points across to domain of interest. A metric known as the error function ($EF$) will be used to facilitate the direct comparison of models tested on the same set of validation data. For model $m$, the error function is defined as:
\begin{equation}
\label{eq:ef}
EF_{m}=\frac{RMSE_{val,m}}{RMSE_{best}}
\end{equation}
where $RMSE_{val,m}$ is the RMSE of model $m$ applied to the validation set, and $RMSE_{best}$ is the lowest error obtained by any model in the comparison for that set of validation data. Consequently, the best modeling method will obtain $EF=1$ while larger values are indicative of inferior solutions.

Two quality metrics will also be used to address the validity of the models in a practical sense. The root mean square error on the validation set of interest $RMSE_{val}$, as well as scaled comparison of the chosen model to an average of the training data. This metric is commonly known as $R^2$ when it is calculated on training data, and is sometimes called $Q^2$ when calculated on validation data. In order to remain consistent, we will adopt the nomenclature $R^2_{val}$ and define the quantity as follows:
\begin{equation*}
R^2_{val} = 1 - \frac{ \sum_{i=1}^{n_{val}} \left( z_i - \hat{z}_i \right)^2}{\sum_{i=1}^{n_{val}} \left( z_i - \bar{z} \right)^2}
\end{equation*}
where $\bar{z}$ is the average of the training response data. Unlike the traditional $R^2$ value, the metric $R^2_{val}$ is not bound below by 0 since a model that is optimal on a training set may generalize worse than the naive model to the validation set. Despite losing the lower bound of 0, $R^2_{val}$ is still bound above by 1 as the error on the validation set can only be driven to 0. In the traditional $R^2$ metric, a value of 1 is often indicative of overfitting; however, measuring this metric on a set of independent validation data alleviates this concern. Instead, a value of 1 is only indicative of a substantially lower error on the validation set than $\bar{z}$, allowing it to be used as a check on model quality.

All simulations were performed on an Intel(R) Core(TM)2 Quad CPU Q9550 at 2.83 GHz. All mixed-integer and nonlinear optimization subproblems were solved using BARON 15.6~\citep{ts:comp:04,s:baron:15.6}. In order to favor a tractable and parsimonious solution, ALAMO employs the use of a heuristic strategy to find an initially appealing linear model. Some problems may terminate in this heuristic phase if the found solution can be proven optimal, others move directly into the mixed-integer optimization of a chosen model fitness metric. In the following computations, ALAMO will utilize the fitness metric Bayesian information criterion, minimizing it by using Formulation~(\ref{eq:bss}). All ALAMO runs utilize Formulation~(\ref{eq:bss}) in order to guarantee the optimal model has been found.

\subsection{Error maximization sampling in ALAMO}

In this section, two experiments are performed to demonstrate the utility of the error maximization sampling methodology in ALAMO. First, ALAMO is given one random point over the training interval, and the EMS methodology is allowed to proceed according to the flowchart in Figure~\ref{fig:alamo} until a tolerance of $\delta=0.0001$ is satisfied. The number of points required to satisfy this tolerance ranges from 14 to 46 across all 150 instances. The models generated by the EMS methodology are then compared to a number of models trained on data obtained from a Latin hypercube design of experiments (LHS). In the first comparison, a Latin hypercube design will be employed for exactly $n_{EMS}$, or the number of points required to satisfy the $\delta$ tolerance for each problem. Next, a Latin hypercube design will be used to generate data sets of the sizes $n \in \{20 , 30 , 40 , 50\}$ (LHS-$N$). Comparing the models generated by the fixed sized training sets to those generated from training sets allowed $n_{EMS}$ sample points serves to directly probe the quality of the solution obtained by the EMS methodology.

\begin{figure*}
\centering
\includegraphics[scale=0.5]{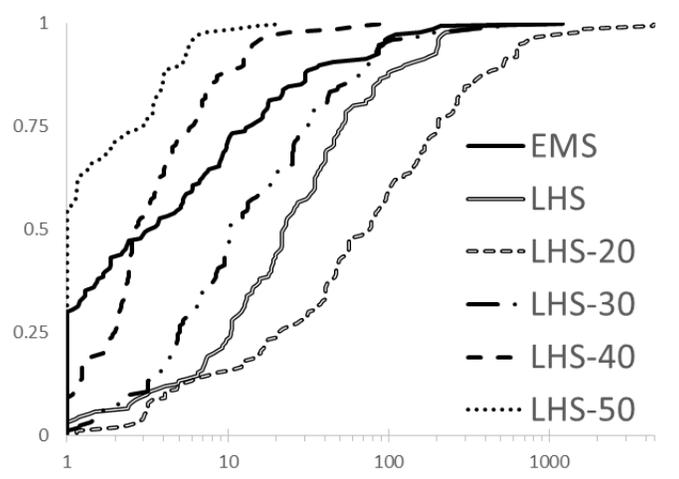}
\caption{Fraction of problems solved vs. error factor}
\label{fig:emsef}
\end{figure*}

\begin{figure*}
\centering
\includegraphics[scale=0.6]{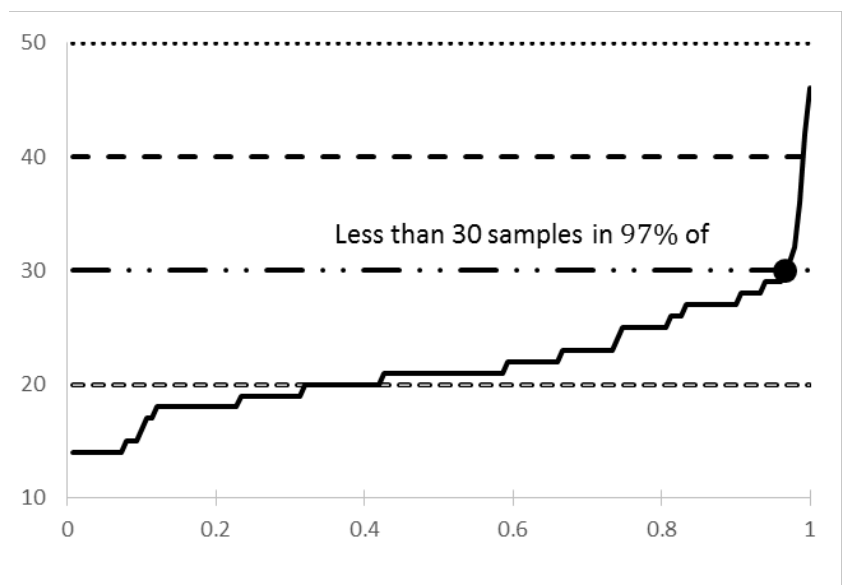}
\caption{Number of data points in training set vs. fraction of problems solve}
\label{fig:emssize}
\end{figure*}

Figure~\ref{fig:emsef} is a Dolan-More{\' e} performance profile~\citep{dm02} that plots the error factor as defined in Equation~(\ref{eq:ef}) on the $x$ axis against the fraction of the 150 problems that achieve the specified error factor. The best performance in this type of plot would be indicated by a vertical line at the origin, meaning one method outperformed all others in all problem instances.  Figure~\ref{fig:emsef} shows that the EMS methodology outperforms a LHS design consisting of the same number of points in the training set, $n_{EMS}$, indicated by the two unbroken lines, in 120 of 150 problems. This demonstrates the efficacy with which the EMS methodology is able to locate valuable new sample points that make a larger contribution towards the final model's predictive ability than those produced by an equivalently sized space-filling method. In the problems in which the LHS methodology outperforms the EMS methodology, two factors are usually present. Despite a difference in error calculated on a validation set, both models result in a high $R^2_{val}$ indicating that the models are near similar in terms of predictive ability. Additionally, in some cases, the EMS methodology requires over 40 points to satisfy the $\delta$ tolerance. In these larger sized examples, the performance of the two methodologies is very similar due to the thorough sampling of the training domain. The exact number of points required by each problem can be seen in Figure~\ref{fig:emssize}, and the general statistics are contained in Table~\ref{tab:psize}. As seen in this table, the EMS methodology requires a maximum of 46 sample points to converge with the specified $\delta -$tolerance, with 97\% of problems requiring less than 30 samples to converge.

% Table generated by Excel2LaTeX from sheet 'erfc work'
\begin{table}[htbp]
  \centering
  \caption{Training set size statistics}
    \begin{tabular}{ccccc}
    \toprule
    \multicolumn{5}{c}{$n_{EMS}$} \\
    \midrule
    Mean  & Median & $\sigma$    & Minimum & Maximum \\
    \midrule
    21.7 & 21    & 4.9 & 14    & 46 \\
    \bottomrule
    \end{tabular}%
  \label{tab:psize}%
\end{table}%

The results from these two methodologies are also compared against LHS designs of a fixed size across all problem instances. At face value, this might seem like an unfair comparison as the static LHS methodologies routinely have access to many more data points than does the EMS methodology. However, the point of this comparison is to demonstrate the quality of the solution obtained by the EMS methodology versus those obtained by space-filling methodologies with access to many more training points on average.  The two unbroken lines in Figure~\ref{fig:emsef} provide a direct comparison of the EMS and LHS curves, while the dashed lines of Figure~\ref{fig:emsef} serve to compare the quality of the solutions obtained by constant-sized LHS designs on much larger training sets.

% Table generated by Excel2LaTeX from sheet 'lhs-ems'
\begin{table*}[htbp]
  \centering
  \caption{Quality metrics calculated on validation set}
    \begin{tabular}{cllllll}
    \toprule
          & \multicolumn{3}{c}{${R^2_{val}}$} & \multicolumn{3}{c}{$\text{RMSE}_{val}$} \\
\cmidrule{2-7}    Sampling method & \multicolumn{1}{c}{Mean} & \multicolumn{1}{c}{$\sigma$} & \multicolumn{1}{c}{Minimum} & \multicolumn{1}{c}{Mean} & \multicolumn{1}{c}{$\sigma$} & \multicolumn{1}{c}{Maximum} \\
    \midrule
    EMS   & 0.994 & 0.02  & 0.906 & 0.0012 & 0.004 & 0.038 \\
    LHS   & 0.988 & 0.04  & 0.547 & 0.0021 & 0.004 & 0.04 \\
    LHS-20 & 0.965 & 0.06  & 0.645 & 0.0063 & 0.01  & 0.083 \\
    LHS-30 & 0.998 & 0.002 & 0.988 & 0.0011 & 0.002 & 0.008 \\
    LHS-40 & 1     & 0     & 1     & 0.002 & 0.0003 & 0.001 \\
    LHS-50 & 1     & 0     & 1     & 0.00009 & 0.0001 & 0.0006 \\
    \bottomrule
    \end{tabular}%
  \label{tab:emsquality}%
\end{table*}%

The models chosen as optimal by ALAMO are compared first by their relative performance on a validation set consisting of 1000 uniformly distributed samples across the batch time domain. Figure~\ref{fig:emsef} demonstrates that the EMS methodology achieves the best performing model in 45 out of 150 problems, only outperformed by the Latin hypercube design of 50 samples (LHS-50) which achieves the best performing model in 83 of 150 samples. Models trained on a design of 30 samples (LHS-30) achieve the best performance on the validation set in only 2 examples. The EMS methodology outperforms the LHS-30 methodology in 103 of 150 samples, despite utilizing less than 30 training points in 147 of 150 problems as seen in Figure~\ref{fig:emssize}.

The number of points in the final training set utilized by the EMS in ALAMO can be seen in Figure~\ref{fig:emssize}. The average number of points in the final training set for all problems is $n_{EMS}=21$; the number of points utilized by the Latin hypercube design remains constant across all problems. This adaptive sampling feature of ALAMO requires 8 iterations in the worst case of 46 data points sampled, and requires more than 40 samples in two problems and more than 30 samples in five problems out of 150.

The astute reader has observed that all sampling schemes compared in this section are implemented in ALAMO.  In other words, ALAMO is capable of working exclusively with a user-specified data set, such as one obtained through LHS.  Additionally, ALAMO is perfectly capable of generating a minimal sampling set through EMS or even combine LHS with EMS.  The trend in the methodologies compared for given static data sets is clear and expected; more data in the training set results in a lower RMSE on a validation set. Therefore, it is important to remember that the solutions obtained by the EMS and LHS methodologies in Figure~\ref{fig:emsef} are not on even footing in terms of the number of points in the training set. Instead, Figure~\ref{fig:emsef} is intended to demonstrate the relative value of the data points obtained by the EMS methodology to those obtained by the space-filling EMS methodologies. Despite being out preformed by Latin hypercube designs consisting of more sampled points, Table~\ref{tab:emsquality} demonstrates the quality of the models produced regardless of the amount or quality of data provided to the wider ALAMO methodology. The lowest $R^2_{val}$ achieved by any experimental design was $R^2_{val}=0.547$ through a 14 points LHS design, and the lowest achieved by the EMS methodology was $R^2_{val}=0.906$. Despite averaging only 21 points in the final data set, the EMS methodology was able to average a $R^2_{val}$ of 0.994 across all 150 problem instances.

\subsection{Constrained regression in ALAMO}

This section will serve to demonstrate the utility of constrained regression in ALAMO. Three experimental settings will be examined: an unconstrained version of ALAMO (UC), a version of ALAMO in which the model output is bound below and above across the training domain (CR), and a bound-extended constrained version of ALAMO that seeks to extend the bound certification beyond the range of the training data (ECR). A non-negative lower bound, and an upper bound constrained by the initial concentration of species A are applied to all constrained regression problems. Training data for all 150 batch-reaction concentration problems will consist of 25 points from a Latin hypercube design of experiments.

\begin{figure*}
\centering
\includegraphics[scale=0.8]{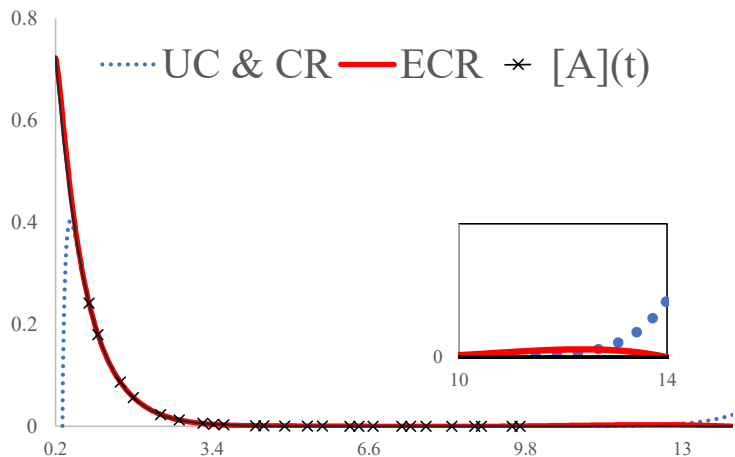}
\caption{Visual comparison of the model produced through bound-extended constrained ALAMO and those produced by constrained and unconstrained ALAMO}
\label{fig:crex}
\end{figure*}

\begin{figure*}
\centering
\includegraphics[scale=0.5]{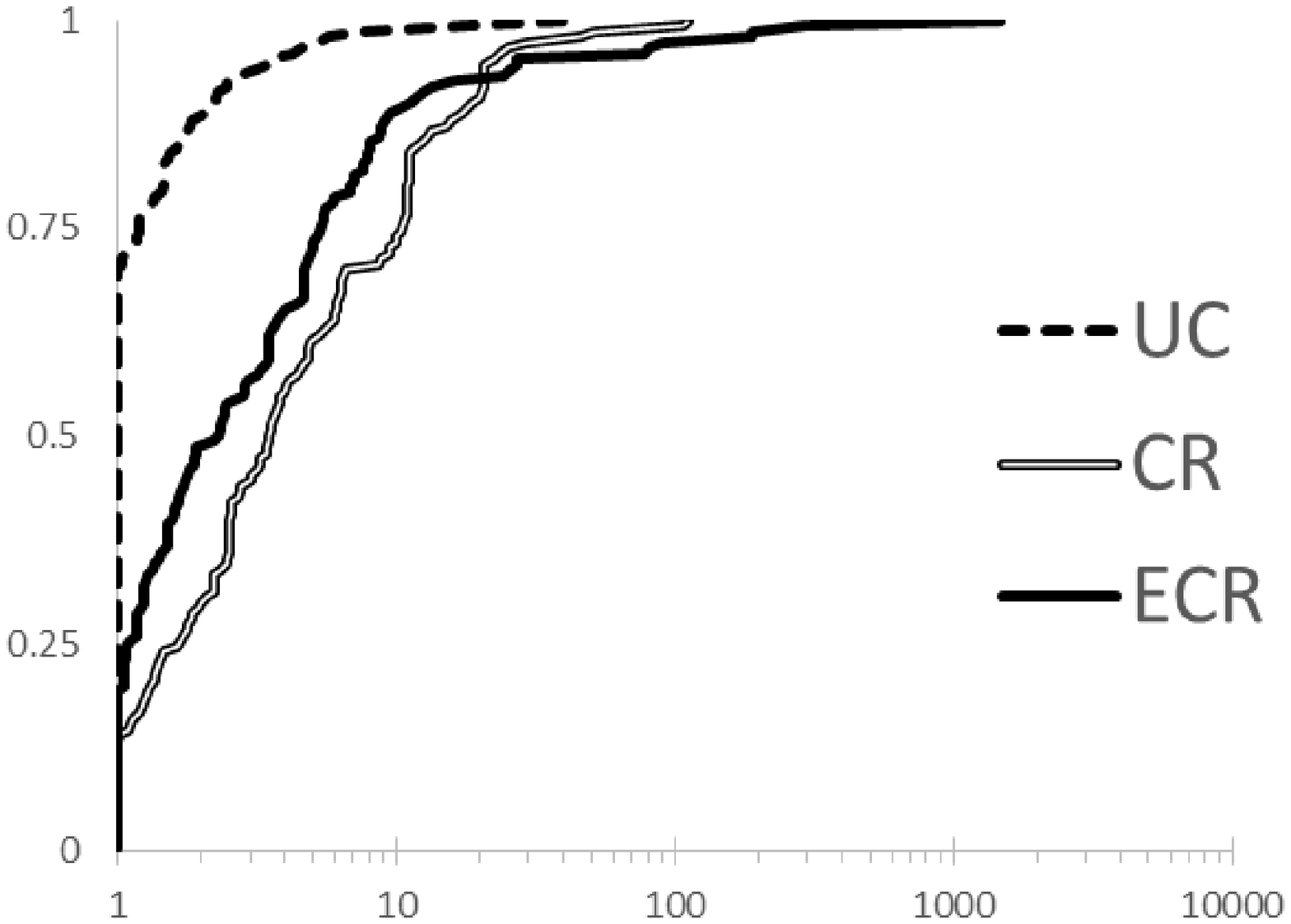}
\caption{Fraction of problems solved vs. error factor, $t \in [0.6 , 10]$}
\label{fig:cdr}
\end{figure*}

\begin{figure*}
\centering
\includegraphics[scale=0.5]{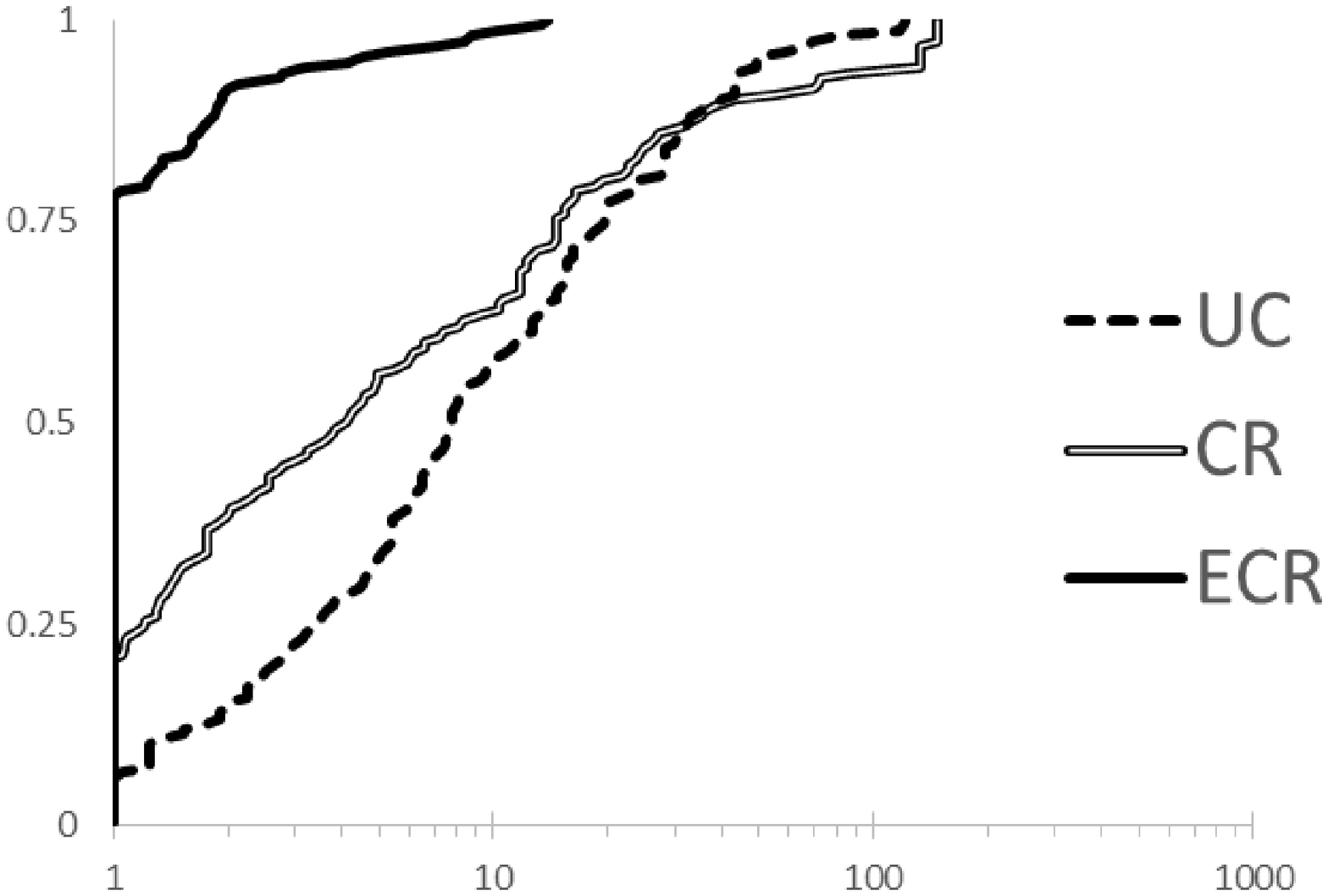}
\caption{Fraction of problems solved vs. error factor, $t \in [0.4 , 0.6)$}
\label{fig:crlb}
\end{figure*}

Figure~\ref{fig:crex} displays the results from the three modeling approaches on the concentration of species A, $[A](t)$, in the case of reactions occurring in parallel. In the illustrative example the CR methodology selected the same model as the UC methodology because it also satisfied the upper and lower bound restrictions over the training domain, $ t \in [ 0.6 , 10 ] $. However, this model did not satisfy the upper and lower bounds over the extended input range, $ t \in [ 0.4 , 14 ] $. Despite the lack of training data present in the range of the extended bounds, a model that satisfies these constraints while maintaining accuracy on the original training domain is identified.
The solution to either constrained regression problem will necessarily have a training error that is the same or larger than the solution to the unconstrained problem. However, this behavior cannot be generalized to any set of independent data, regardless of whether it is present inside or outside of the training domain. As such, the models produced by both CR and ECR may result in higher validation errors, even if the bounds they seek to maintain are appropriate and satisfied. This behavior, namely a decrease in validation error in the region outside the training set where CR is used to guarantee bounds at the expense of an increase in validation error in the original training domain, is observed throughout the synthetic reaction problem set, as seen in Figure~\ref{fig:cdr} and Figure~\ref{fig:crlb}.

In the training domain, $ t \in [ 0.6 , 10 ] $, as seen in Figure~\ref{fig:cdr}, models generated by UC have the lowest validation error in 103 problems, followed by ECR with 29, and CR performing best on 20 problems. Table~\ref{tab:crq} displays the quality metrics $\text{RMSE}_{val}$ and $R^2_{val}$ as applied to  validation data obtained in the training domain. As expected, UC obtains the lowest average $\text{RMSE}_{val}$ and highest average $R^2_{val}$ values. Moreover, the high absolute values of $R^2_{val}$ in this domain are indicative of high quality solutions in almost all cases.

Validation data with $t$ values higher than those in the training domain demonstrate the utility of both constrained regression approaches. When the constrained regression feature is utilized to extend the upper bound of the time domain from 10 to 14, the UC approach and the CR approach perform best on 59 and 57 problems respectively, with the ECR approach achieving the lowest validation error on the remaining 36. Although it may seem counter intuitive that the models that obey physical constraints achieve a lower validation error, the satisfaction of the upper and lower bound constraints comes at the expense of a higher residual elsewhere in the model. This can be seen in a wide number of examples in~\cite{csm:15} utilizing training data.

The utility of the ECR methodology can be seen most clearly in Figure~\ref{fig:crlb}, which shows it outperforming the other methodologies in 117 problems. The CR methodology performs best in 29 problems and the unconstrained models produced using the UC methodology perform best in the remaining 9. The quality metrics in Table~\ref{tab:crq} demonstrate the difficulty of generating an accurate model over this domain with the chosen 13 nonlinear transformations applied to the input variable $t$.  However, Table~\ref{tab:crq} also demonstrates the ability of both the CR approach and ECR approach to generate accurate models in domains in which they have not been trained while satisfying physical constraints.

\section{Conclusions}
\label{concs}

This paper presented ALAMO, a computational methodology and software developed to address the fundamental problem of leaning algebraic functions from data sets. A model fitness metric of choice, in this paper the Bayesian information criterion, is used to balance the bias-variance trade-off incumbent upon finding the best subset of a large number of explicit non-linear transformations of the process inputs used to construct a linear surrogate model. The model can be refined, as additional data are obtained in an adaptive fashion through the use of derivative-free optimization and the EMS methodology. This adaptive sampling methodology is able to make efficient use of small amounts of data, and outperforms space-filling models with larger training sets.  An additional technique, referred to as constrained regression, where the model response is controlled through constraints on the regression parameters can be used additionally to enforce first-principles-based constraints on the model response. Forcing these models to obey physical constraints through constrained regression can result in models that are able to accurately predict data outside of the range in which the model was trained. The utility of the ALAMO methodology to generate simple algebraic models was demonstrated on a number of problems, and the additional performance imparted by using error maximization sampling as well as a priori domain knowledge through constrained regression was demonstrated.

\section{Acknowledgments}

As part of the National Energy Technology Laboratory's Regional University Alliance (NETL-RUA), a collaborative initiative of the NETL, this technical effort was performed under project 1042568, as part of the Institute for the Design of Advanced Energy Systems. This report was prepared as an account of work sponsored by an agency of the United States Government. Neither the United States Government nor any agency thereof, nor any of their employees, makes any warranty, express or implied, or assumes any legal liability or responsibility for the accuracy, completeness, or usefulness of any information, apparatus, product, or process disclosed, or represents that its use would not infringe privately owned rights. Reference herein to any specific commercial product, process, or service by trade name, trademark, manufacturer, or otherwise does not necessarily constitute or imply its endorsement, recommendation, or favoring by the United States Government or any agency thereof. The views and opinions of authors expressed herein do not necessarily state or reflect those of the United States Government or any agency thereof.

% Table generated by Excel2LaTeX from sheet 'cr r2'
\begin{landscape}
\begin{table}[htbp]
  \centering
  \caption{Quality metrics calculated on validation set}
    \begin{tabular}{ccccccccccc}
    \toprule
          &       & \multicolumn{3}{c}{\textbf{Lower extended domain}} & \multicolumn{3}{c}{\textbf{Training domain}} & \multicolumn{3}{c}{\textbf{Upper extended domain}} \\
\cmidrule{3-11}          &       & UC    & CR    & ECR   & UC    & CR    & ECR   & UC    & CR    & ECR \\
\cmidrule{3-11}    \multirow{4}[2]{*}{$ \text{RMSE}_{val} $} & Mean  & 1.95  & 1.98  & 0.27  & 1.1E-03 & 3.0E-03 & 3.1E-02 & 4.4E-03 & 7.6E-03 & 4.1E-02 \\
          & $\sigma$ & 3.1   & 3.6   & 0.49  & 1.7E-03 & 4.3E-03 & 1.6E-01 & 7.1E-03 & 2.0E-02 & 1.7E-01 \\
          & Minimum & 0.06  & 0.01  & 0.017 & 2.2E-05 & 9.4E-05 & 1.1E-04 & 4.4E-05 & 4.3E-07 & 9.3E-10 \\
          & Maximum & 13.6  & 23    & 3.39  & 7.7E-03 & 2.8E-02 & 9.7E-01 & 3.9E-02 & 1.5E-01 & 1.0E+00 \\
    \midrule
    \multirow{4}[2]{*}{$ R^2_{val}$} & Mean  & -111  & -102  & -5.2  & 0.99  & 0.99  & 0.93  & 0.99  & 0.99  & 0.91 \\
          & $\sigma$ & 326   & 382   & 24.8  & 0.0001 & 0.001 & 0.4   & 0.001 & 0.03  & 0.46 \\
          & Min & -2180 & -3810 & -203  & 0.999 & 0.995 & -1.77 & 0.98  & 0.73  & -2 \\
          & Max & 0.983 & 0.996 & 0.998 & 1     & 1     & 1     & 1     & 1     & 1 \\
    \bottomrule
    \end{tabular}%
  \label{tab:crq}%
\end{table}%
\end{landscape}

\end{singlespacing}

\end{document}